\documentclass[10pt,journal,compsoc]{IEEEtran}

\ifCLASSOPTIONcompsoc
  \usepackage[nocompress]{cite}
\else
  \usepackage{cite}
\fi

\ifCLASSINFOpdf
\else
\fi

\usepackage{hyperref}
\usepackage{array}
\usepackage{url}
\usepackage{algorithm,algorithmic}
\usepackage{calc}
\usepackage{graphicx}
\usepackage{enumitem}
\usepackage{caption}
\usepackage{subcaption}
\usepackage{bbm, dsfont}
\usepackage{caption}
\usepackage{multirow}
\usepackage{amssymb,amsmath}
\usepackage{array,booktabs,ragged2e}
\usepackage{balance}
\newcolumntype{L}[1]{>{\raggedright\let\newline\\\arraybackslash\hspace{0pt}}m{#1}}
\newcolumntype{C}[1]{>{\centering\let\newline\\\arraybackslash\hspace{0pt}}m{#1}}
\newcolumntype{R}[1]{>{\raggedleft\let\newline\\\arraybackslash\hspace{0pt}}m{#1}}

\newcommand{\argmin}{\arg\!\min}
\newcommand{\argmax}{\arg\!\max}

\usepackage{color,soul}
\soulregister\cite7
\soulregister\ref7
\soulregister\pageref7

\begin{document}
%
\title{ATD: Anomalous Topic Discovery in High Dimensional Discrete Data}


\author{Hossein~Soleimani,
    and~David~J.~Miller,~\IEEEmembership{Senior~Member,~IEEE}
\IEEEcompsocitemizethanks{\IEEEcompsocthanksitem H. Soleimani and D. J. Miller are with the School of Electrical Engineering and Computer Science, Pennsylvania State University, University Park, PA, 16802.\protect\\
E-mails: hsoleimani@psu.edu, djmiller@engr.psu.edu}
}


%


\IEEEtitleabstractindextext{%
\begin{abstract}
We propose an algorithm for detecting patterns exhibited by anomalous clusters in high dimensional discrete data. Unlike most anomaly detection (AD) methods, which detect individual anomalies, our proposed method detects groups (\textit{clusters}) of anomalies; i.e. sets of points which collectively exhibit abnormal patterns. In many applications this can lead to better understanding of the nature of the  atypical behavior and to identifying the sources of the anomalies. Moreover, we consider the case where the atypical patterns exhibit on only a small (salient) subset of the very high dimensional feature space. Individual AD techniques and techniques that detect anomalies using all the features typically fail to detect such anomalies, but our method can detect such instances collectively, discover the shared anomalous patterns exhibited by them, and identify the subsets of salient features. In this paper, we focus on detecting anomalous topics in a batch of text documents, developing our algorithm based on topic models. Results of our experiments show that our method can accurately detect anomalous topics and salient features (words) under each such topic in a synthetic data set and two real-world text corpora and achieves better performance compared to both standard group AD and individual AD techniques. All required code to reproduce our experiments is available from \href{https://github.com/hsoleimani/ATD}{https://github.com/hsoleimani/ATD}.
\end{abstract}

\begin{IEEEkeywords}
Anomaly Detection, Pattern Detection, Topic Models, Topic Discovery.
\end{IEEEkeywords}}

\maketitle



\IEEEdisplaynontitleabstractindextext

%
\IEEEpeerreviewmaketitle

\ifCLASSOPTIONcompsoc
\IEEEraisesectionheading{\section{Introduction}\label{sec:introduction}}
\else
\section{Introduction}
\label{sec:introduction}
\fi

\IEEEPARstart{A}{nomaly} detection (AD) is the problem of identifying items or patterns which do not conform to normal or expected behavior \cite{HodgeVictoriaJ.2004},\cite{Chandola2009}. Anomaly detection techniques have been widely used e.g. to detect credit card fraud \cite{Srivastava2008}, insurance fraud \cite{Major2002}, and network intrusions \cite{Wang2004},\cite{Kocak2014}.

AD techniques typically detect individual sample anomalies. In this work, however, we focus on detecting abnormal \textit{patterns} exhibited by anomalous groups ({\it clusters}) of samples. An anomalous cluster is a set of data samples which manifest similar patterns of atypicality. Each of the samples in such a cluster may not be highly atypical by itself, but, when considered collectively, the cluster demonstrates a distinct pattern which is significantly different from expected (normal) behavior. In this paper, we propose a framework to detect such groups of anomalies and the atypical patterns they exhibit. Moreover, we consider the case where the anomalous pattern may manifest on only a small subset of the features, not on the entire feature space; {\it i.e.} samples in the anomalous cluster may be far apart from each other measured on the full feature space, but on a subset of the feature space (the salient features), they exhibit a similar pattern of abnormality. In addition to detecting atypical clusters, our proposed method identifies each cluster's salient feature subset.

In some cases, no prior knowledge about normal behavior is available, and the goal is to detect anomalies (outliers) in a single data set consisting of normal and possibly abnormal instances, without any annotation of which samples are normal. More typically, and as we assume here, there is a collection of normal data which sufficiently characterizes normal behavior. In the training phase, we use this data to build a (null) model. Then, in the detection phase, this model is used as a reference to help detect (possible) clusters of anomalous patterns in a different (test batch) data set. 

\noindent {\it Illustrative Applications:}

Our proposed framework has significant applications in a variety of domains. For instance, consider an open repository of scientific or business related articles. A company may try to post articles on this repository to promote its products or services. However, to avoid being easily detected by normal advertisement blocker services, the articles are written in such a way that they match the normal articles on that repository in form and content. Only a small part of these advertising articles promote the company's services. In this case, we can identify that company's infiltration by detecting clusters of such articles. In order to do so, we first use a sub-collection of normal articles from that repository as our training set to learn the normal topics (null model). Then, using that null model, our algorithm detects clusters of such advertising articles within the full repository, the anomalous topic of each cluster (the product or service they promote), and the keywords representing that topic. 

Some other potentially important applications of our framework are: detecting similar patterns in malware and spyware (that were uploaded to a public software tool repository) to identify sources of attacks; studying patterns of anomalies in consumer behavior to discover emerging consumer trends; finding shared patterns of tax avoidance to reveal loopholes in the law; and detecting organized malicious activities in social media.

\noindent {\it Formal Problem Statement:}

\noindent We formally define our problem as follows.
\newline
{\bf Given:} A training set consisting solely of normal data to be used in learning a null model.
\newline
{\bf Given:} A test set of samples, some of which may be normal and some of which may be abnormal.  Furthermore, the abnormal sample subset may consist of {\it clusters} of samples, with each cluster distinctively characterized
by the fact that its samples exhibit anomalous behavior (relative to the null model) on the same {\it low-dimensional} subset of the full (high-dimensional) feature space.
\newline
{\bf Objective:} Detect the clusters of anomalous samples in the test batch and identify the salient feature 
subset for each such cluster.

For example, the test batch could consist of 10,000 samples, each defined on a feature space
$\underline{X} \in {\cal R}^{1000}$.  There could be two anomalous clusters in the test batch, with one cluster
consisting of 50 samples, all of which exhibit anomalous behavior with respect to the (very low) four-dimensional
feature subspace $(X_{17},X_{32},X_{241},X_{379})$.  Another anomalous cluster could consist of 100 samples, each
exhibiting anomalous behavior with respect to the six-dimensional feature subspace $(X_{14},X_{79},X_{256},X_{439},X_{597},X_{801})$.  Note that these two clusters each exhibit {\it normal} behavior on the (very large) remaining subset 
of the full feature space.

The problem of detecting clusters of data points which exhibit similar anomalous patterns is sometimes referred to as {\it group anomaly detection} \cite{Xiong2011a},\cite{Yu},\cite{Muandet2013}.  We will synonymously refer to {\it detecting clusters of anomalies} and {\it group anomaly detection}. Unlike these previous works, the focus here is on group anomaly detection in very high-dimensional data domains,
where the samples in a group 
are expected to manifest their anomalies on a (the same) low-dimensional ({\it a priori} unknown) subset of the high-dimensional
feature space. Thus, our approach requires jointly detecting these clusters of samples and their
(in general, low-dimensional) salient feature subsets. 

Our anomalous cluster detection approach consists of two fundamental steps repeatedly applied to the test batch: i) determining the best current candidate anomalous cluster; ii) determining whether this candidate cluster is anomalous. Note that we do not presume that any anomalous clusters actually exist in the test data. In this paper, we propose statistical tests to accomplish both these steps; {\it i.e.}, to determine which samples significantly belong to the best current cluster candidate and to test whether the candidate exhibits a statistically significant degree of atypicality relative to the null model.

Our proposed framework can be applied generally, to both continuous and discrete valued data. However, in this paper, we focus on detecting atypical patterns (topics) in text documents, a domain with a very high-dimensional (bag-of-words) feature space. Anomalous topic discovery (ATD) for document databases represents a challenging domain due to the high feature dimensionality, with \textit{many} candidate low-dimensional subspaces that may exhibit anomalous patterns. We develop our proposed framework focusing on topic models \cite{Blei2003},\cite{Blei2010a}. Topic models have been used in modeling different types of data such as images and text documents. In this paper, we use the formulation of topic models primarily developed for modeling text documents, based on a multinomial distribution model for each topic \cite{Blei2003}.  

Topic models are a class of statistical models often used for discovering latent patterns (topics) in a collection of text documents. Each topic specifies a pattern of words; i.e. words that appear more or less frequently than others under that topic. A simple and yet widely popular topic model is Latent Dirichlet Allocation (LDA) \cite{Blei2003}, which posits document-specific mixing proportions over the topics, with each topic a multinomial distribution over the given vocabulary.

LDA in its basic structure is a parameter-rich model, which, when applied to high-dimensional problems such as text documents may result in poor generalization and semantically uninterpretable topics. \cite{Soleimani2014} extends LDA by proposing Parsimonious Topic Models (PTM). PTM controls the number of free parameters in the model by balancing model complexity and goodness of fit to the data set used for learning the model. \cite{Soleimani2014} hypothesizes that, under each topic, only a modest number of words have topic-specific characteristics (salient words), which warrant their own probability parameters, while the rest of the words can be described by a universal shared model across all topics. Moreover, PTM proposes that only a sparse subset of topics are present in each document, with the rest of the topics having zero proportions. \cite{Soleimani2014} shows that PTM achieves better generalization accuracy (classification and test set log-likelihood) than LDA evaluated on multiple text corpora.

\cite{Soleimani2014} optimizes an objective function, a Bayesian Information Criterion (BIC), specifically derived for the PTM structure, to jointly learn the structure of the model (the set of topic-specific words under each topic and the set of active topics in each document) and to estimate the model parameters (word probabilities and active topic proportions). Moreover, the PTM objective function is also optimized with respect to (thus estimating) the number of topics (model order) present in the corpus. 

We choose PTM over LDA as the topic model for our ATD algorithm for a number of reasons.  First, because PTM typically achieves better generalization accuracy than LDA and it automatically estimates the number of normal topics, unlike LDA, which requires this number to be set by a user \cite{Soleimani2014}. Note that model order selection is a crucial step in anomalous topic discovery. Specifically, since significance of any anomalous topic will be measured with respect to the null model (normal topics), either under {\it or} overfitting the null can lead to false discovery of anomalous clusters due, respectively, to limited modeling power or to poor generalization. Moreover, PTM, unlike LDA, identifies a highly sparse set of topic-specific (salient) words for each topic. This makes PTM a natural fit for our ATD algorithm as we assume that the anomalous topics manifest on a very low-dimensional subspace of the full word space. PTM, with its sparse topic representation, is expected to have an inherent performance advantage over LDA, which uses all the words in the dictionary to define topics.  In fact, this is supported
by our experimental results in the sequel.

Our anomalous topic discovery (ATD) algorithm consists of two main parts: First, in the training step, we learn PTM as our null model $\mathcal{M}_0$, with $M$ its estimated number of topics. The {\textit{null hypothesis}} is that all documents in the test set were generated by the null model. Second, in the detection phase, under the \textit{alternative hypothesis}, we posit that a cluster of documents in the test set may contain an additional topic. Accordingly, we build an alternative model $\mathcal{M}_1$ with $M+1$ topics by simply adding one topic to the null model. Then, in the spirit of a generalized likelihood ratio test, we seek the best \textit{candidate} anomalous cluster by, alternately, learning the parameters of the new topic and choosing the documents from the test set for which the new topic has significant presence, until a convergence criterion is met. Finally, we measure statistical significance of the candidate cluster. If the cluster is significant, we detect it, remove all its documents from the test set, and repeat this detection process until no further significant topics are discovered. That is, we detect anomalous clusters in the test set one by one. Note that the new topic represents the anomalous pattern in each cluster and the set of topic-specific words under that topic are the salient features of that pattern.

We apply non-parametric bootstrap testing \cite{Efron1979} both to determine i) whether a document belongs to a candidate anomalous cluster and ii) if a candidate cluster is significantly anomalous. For the first task, we compare the empirical topic proportion of the new topic in the candidate document with that of a set of normal bootstrap documents. Similarly, for measuring significance of a cluster, we compute the ratio of the candidate cluster's likelihood under the alternative and null models, comparing it with normal bootstrap clusters, and compute an empirical p-value. We call a candidate cluster anomalous only if the empirical p-value is lower than a pre-set significance level.

\subsection{Related work}
In this section, we review some previous works on group anomaly detection. \cite{Xiong2011a} proposes a Mixture of Gaussian Mixture Models (MGMM) for group anomaly detection. \cite{Xiong2011a} assumes each data point belongs to one group and that all points in a group are modeled by the group's Gaussian mixture model. Mixing proportions of the mixture model for each group, however, are not freely estimated, but rather, in a hierarchical way, are selected from a limited set of $T$ possible mixing proportion ``types" (genres). These types represent the normal behaviors. A test group is called anomalous if it has low likelihood under the normal types. This idea is then extended to Flexible Genre Models (FGM) in \cite{Xiong2011} by treating mixing proportions as random variables which are conditioned on possible normal genres. One significant shortcoming of these methods is that they assume that the group membership for every data point is known {\it a priori}.  Since this information is not available in general, one must in practice perform hard clustering of the data into groups prior to applying FGM or MGMM. Such clustering, working in the full (high-dimensional) feature space, may be highly inaccurate when the anomalous pattern lies on a low-dimensional feature subspace. Another major issue with these methods is that they do not provide any significance test for group anomalies -- they simply declare a candidate cluster anomalous if it is among the top $K\%$ of clusters with highest anomaly scores or if its anomaly score is higher than a pre-set threshold value. Note that the proper choice of such thresholds is problem-dependent -- a poorly chosen threshold may lead either to a high false detection rate or to weak detection power.

\cite{Yu} addresses the first issue by presenting a method, specifically for network analysis, for jointly detecting groups of similar nodes and computing anomaly scores for the discovered groups. Nevertheless, unlike our method, \cite{Yu} does not have an algorithmic procedure for discovering ``hard" anomalous clusters one by one -- some post-processing effort is required to hard-assign each data point to the cluster with highest membership degree. Moreover, \cite{Yu} does not provide any statistical significance testing and relies on choosing an appropriate threshold for detecting anomalous clusters. 

\cite{Muandet2013} follows a discriminative approach to group anomaly detection and generalizes the idea of one-class support vector machines to a space of probability measures, proposing one-class support measure machines. Groups in this method are represented as probability distributions which are mapped into a reproducing kernel Hilbert space using kernel methods. Similar to MGMM, this method requires hard-clustering of the data prior to detecting any anomalous group. 

\cite{Wong2002} proposes a rule-based anomalous pattern discovery algorithm for detecting disease outbreaks. Anomalous patterns in this method are characterized by first or second order ``rules''. Each rule is simply a set of possible values that a subset of categorical features take on. Significance of each rule is measured by comparing occurrence frequency of each rule in the test set relative to the training set by conducting Fisher's exact test and a randomization test. This idea is then extended in \cite{Wong2003a}, which uses Bayesian networks to measure relative significance of each rule.  \cite{Das2008} uses a similar procedure, but first detects individual anomalous points and then searches for possible patterns among them.
These methods do provide statistical testing procedures to measure significance of each cluster. They can also (for very low dimensional problems) detect salient features for each cluster. But, unlike our method, they do not provide an optimization algorithm for jointly detecting clusters and their associated low-dimensional anomalous patterns. This, in particular, makes these methods less suitable for high dimensional domains (such as text documents).

\cite{McFowland2013} proposes Fast Generalized Subset Scan (FGSS) to detect anomalous patterns in categorical data sets. Unlike many other methods, FGSS provides an algorithm for constructing anomalous clusters by jointly searching over subsets of data instances and subsets of anomalous attributes. FGSS has better scaling characteristics than \cite{Das2008} and \cite{Wong2003a} and, thus, can detect anomalous patterns which lie on higher dimensional feature spaces. However, FGSS requires computing a p-value for each feature of every sample based on a Bayesian network learned on the training set. Learning Bayesian networks may not be practically feasible for high-dimensional problems such as text documents where there may be tens of thousands of features. Moreover, FGSS can only detect a subset of anomalous features for each cluster -- unlike our method, FGSS does not provide a \textit{model} for the common pattern of anomalies exhibited by the cluster.

A somewhat related problem to anomalous topic detection in text documents is the problem of Topic Detection and Tracking (TDT) in the information retrieval literature. The main focus of TDT is tracking topics and detecting new events in a temporally ordered stream of articles \cite{Allan1998a},\cite{Dai2010}. TDT methods, in general, fall into the category of clustering evolving data streams and single shot clustering, and extensively rely on the temporal location of each document and other related meta-data. In fact, even in offline (batch) TDT, time is a central part of the analysis \cite{He2010}. Our method, on the other hand, considers a batch of documents (bag-of-word objects) and discovers anomalous topics solely based on contents of the documents -- our method does not exploit any temporal information about the documents.

\textit{Contributions of this paper:} Compared to previous works, our main contributions are:
\begin{enumerate}
\item We propose an algorithm to \textit{jointly} learn and detect anomalous clusters and the (low-dimensional) anomalous patterns that they exhibit. Most prior works require separate procedures for clustering the data and for measuring the degree of anomaly.
\item Our algorithm does not require any user setting of thresholds on score values to detect anomalous clusters. Instead, we propose tests to measure statistical significance of any candidate anomalous cluster, compared to normal clusters. 

Our method still requires setting thresholds on p-values for determining statistical significance of candidate anomalous clusters. However, setting thresholds on p-values is a common practice in statistical hypothesis testing, and is much easier and more interpretable (with respect to controlling false positive rates) than setting thresholds on more general score functions, which do not have a probabilistic interpretation.
\item Our algorithm is able to discover anomalous patterns that may lie on a very low-dimensional subspace of a high-dimensional feature space, thus identifying the salient features of the anomalous cluster.
\item Our approach non-trivially adapts PTM \cite{Soleimani2014} from document clustering to detecting {\it clusters of anomalies on low-dimensional feature subspaces}.
\end{enumerate}

The rest of the paper is organized as follows: Section \ref{PTMsection} reviews PTM. In section \ref{ATDsection}, we present our anomalous topic discovery algorithm. Section \ref{Expsection} gives our experimental results. We summarize the experimental results and further discuss our algorithm in section \ref{disc_section}. Finally, in section \ref{Concsection}, we give concluding remarks.

\section{Parsimonious Topic Model}
\label{PTMsection}
In this section, we review the Parsimonious Topic Model (PTM) \cite{Soleimani2014}, the model that we use to learn normal topics (the null model) on the training set and later to discover anomalous topics in the test set batch.

We assume that the training corpus, $\mathcal{D}$, is a collection of $D$ normal documents indexed by $d \in \{1,2,...,D\}$. There are, in total, $N$ unique words in the dictionary, indexed by $n \in \{1,2,...,N\}$. Document $d$ consists of $L_d$ words $\{w_{1d},...,w_{L_dd}\}$ where $w_{id} \in \{1,2,...,N\}$ $i=1,2,...,L_d$. Our goal in the training phase is to extract $M$  normal patterns (``topics'') from the training documents (and, to jointly estimate $M$). We index each topic by $j \in \{1,2,...,M\}$.

Each topic is a multinomial distribution over all words in the dictionary, $(\beta_{j1},...,\beta_{jN})$ where $\sum_{n=1}^{N}\beta_{jn} = 1~\forall j$. In a simple topic model such as LDA, each word has a topic-specific probability parameter under every topic. For a typical corpus, this can amount to hundreds of thousands of free parameters and hence may result in models with poor generalization performance. This is also in contrast to our understanding of human languages, wherein only a small subset of words should have context-dependent characteristics, with the rest used with relatively the same frequencies under different topics. PTM proposes the concepts of {\it topic-specific} and {\it shared} words. Under each topic, a subset of words are identified as topic-specific, having their own probability parameters, while the rest of the words are modeled by a universal shared model. 
In this way, PTM gives a sparse representation of each topic.
More specifically, the probability of word $n$ under topic $j$ is $\beta^{u_{jn}}_{jn}\beta^{1-u_{jn}}_{0n}$, $u_{jn}$ a binary switch variable which specifies if word $n$ under topic $j$ is topic-specific $(u_{jn}=1)$ with parameter $\beta_{jn}$ or shared $(u_{jn}=0)$, with globally shared parameter $\beta_{0n}$. Note that for any given configuration of $u$ switches and probability parameters, each topic must be a valid probability mass function (pmf), {\it i.e.} $\sum_{n=1}^N\beta^{u_{jn}}_{jn}\beta^{1-u_{jn}}_{0n}=1, j=1,...,M$.

Under LDA, each document is assumed to be generated by a mixture model with $M$ (topical) components. The mixing proportions, called topic proportions, for document $d$ are $(\theta_{1d},...,\theta_{Md})$, where $\sum_{j=1}^M \theta_{jd} = 1 ~\forall d$. Under LDA, in general, each topic is present with non-zero proportion in every document (even if some of these proportions are very small). However, in practice, each document is expected to only contain a modest set of topics, with the rest of the topics having \textit{no} presence. PTM encodes sparsity in topic proportions via a binary switch variable $v_{jd}$, which specifies whether topic $j$ is absent ($v_{jd}=0$) (has zero proportion) or present ($v_{jd}=1$), with non-zero proportion $\theta_{jd}$. Given the switch configuration and topic proportion parameters, we have a multinomial distribution for every document, {\it i.e.} $\sum_{j=1}^M\theta_{jd}v_{jd}=1~\forall d$. 
Note that in LDA, with the proper choice of the hyper-parameter of the Dirichlet prior on topic proportions, posterior estimates for some topic proportions may be much smaller than others. However, as shown experimentally in \cite{Soleimani2014}, this ``approximate sparsity'' in \textit{posterior} topic proportions does not have the same impact on model performance as the \textit{structural sparsity} (parsimony) in topic proportions in PTM (wherein many of the topic proportions in a document are \textit{identically} zero) -- compared to LDA, the parsimony in topic proportions in PTM along with the parsimony in word probabilities result in more interpretable topics and better generalization performance on test documents \cite{Soleimani2014}. An analogous parsimonious treatment of topic proportions in LDA would require ``mapping'' the original Dirichlet prior on topic proportions in each document from an $(M-1)$-dimensional simplex to an \textit{a priori} unknown lower $(M_d-1)$-dimensional simplex, where $M_d$ is the number of topics present in document $d$.

Note that the subsets of topic-specific and shared words as well as the subset of active topics in each document are {\it a priori} unknown and need to be determined, along with all other model parameters. That is, training of PTM involves determining two types of parameters: i) structural parameters $\mathcal{H} = \{M,\{u\},\{v\}\}$: $u$ switches, $v$ switches, and the number of topics $M$; ii) model parameters ${\Theta}=\{\{\beta\},\{\theta\}\}$: shared and topic-specific word probabilities as well as topic proportions. These two classes of parameters collectively specify a PTM model $\mathcal{M}=\{\mathcal{H},{\Theta}\}$.

PTM assumes the following generative process for each document $d$ in a corpus $\mathcal{D}$.
\begin{itemize}[leftmargin=*]
\item For each word $i=1,\ldots,L_d$
\begin{enumerate}
\item Select a topic represented by the M-dimensional binary random vector $z_{id}\sim \text{Multinomial}(\theta_{1d}v_{1d},...,\theta_{Md}v_{Md})$. 
\item Given the selected topic, randomly select the $i$-th word based on the topic's
pmf over the word space $p(w_{id}|z_{id},\beta,u),~w_{id}\in\{1,2,...,N\}$,
\end{enumerate}
\end{itemize}
where $p(w_{id}|z^{(j)}_{id}=1,\beta,u) = {\beta^{u_{j{w_{id}}}}_{jw_{id}}}{\beta^{1-u_{j{w_{id}}}}_{0w_{id}}}$.

Based on this generative process, the likelihood of the corpus $\mathcal{D}$ under PTM is:
\begin{equation}
p(\mathcal{D}|\mathcal{H},\Theta)=\prod_{d=1}^{D}\prod_{i=1}^{L_d}\sum_{j=1}^{M}\big[\theta_{jd}v_{jd}p(w_{id}|z^{(j)}_{id}=1,\beta,u)\big].
\label{ptmlkh}
\end{equation}
In principle, both the structural and model parameters could be chosen to maximize the likelihood function (\ref{ptmlkh}). However, na\"{\i}vely maximizing (\ref{ptmlkh}) will yield $u_{jn}=1 ~\forall j,n$ and $v_{jd}=1,~\forall j,d$, as such a model has the greatest number of free parameters, which can be used to increase (\ref{ptmlkh}); however these choices will tend to {\it overfit} the model to the
limited training data set. Instead, PTM optimizes a \textit{penalized} likelihood function, exploiting 
vast degrees of freedom in defining a sparse (parsimonious) model.  Such models 
achieve better performance on unseen (test) documents than LDA \cite{Soleimani2014}. To objectively achieve this goal, \cite{Soleimani2014} derives a Bayesian Information Criterion (BIC) \cite{Schwarz1978} for PTM by applying Laplace's approximation to the negative logarithm of the marginal likelihood function and defining proper priors on the model structure to promote sparsity. The BIC objective function for PTM is \cite{Soleimani2014}:
\begin{equation}
BIC(\mathcal{D},\mathcal{H},\Theta) = Cost(\mathcal{H},\mathcal{D}) - \log(p(\mathcal{D}|\mathcal{H},{\Theta})),
\label{bic}
\end{equation}
where
\begin{flalign}
\label{biccost}
Cost(\mathcal{H}, \mathcal{D})&=D\log(M)+\sum_{d=1}^{D}{\log\binom{M}{M_d}}\\
&+MNH(\frac{\bar{N}}{N})\log(2)-\frac{1}{2}\log(MN) \nonumber\\
&+\frac{1}{2}\sum_{d=1}^{D}{(M_d-1)\log(\frac{L_d}{2\pi})}+\frac{1}{2}\sum_{j=1}^{M}{N_j\log(\frac{\bar{L}_j}{2\pi})}.\nonumber
\end{flalign}

Here, $M_d = \sum_{j=1}^Mv_{jd}$ is the number of active topics in document $d$, $N_j = \sum_{n=1}^Nu_{jn}$ is the number of topic-specific words in topic $j$, $\bar{L}_j = \sum_{d=1}^D L_dv_{jd}$ is the total length of documents in which topic $j$ is active, and  $\bar{N}=\frac{1}{M}\sum_{j=1}^{M}N_j$ is the average number of topic-specific words across all topics. Also, $H(\frac{\bar{N}}{N})$ is Shannon's entropy for a Bernoulli random variable with probability $\frac{\bar{N}}{N}$.

BIC consists of two main terms: i) The negative log-likelihood term which depends on both the structure and model parameters; ii) The cost $Cost(\mathcal{H}, \mathcal{D})$ which penalizes model complexity and promotes sparsity. $Cost(\mathcal{H}, \mathcal{D})$ only depends on the structural parameters. By minimizing (\ref{bic}), PTM jointly estimates both the structure and model parameters. 

Note that unlike the na\"{\i}ve form of BIC \cite{Schwarz1978} in which \textit{all} model parameters are penalized equally (proportional to $\log(\text{sample size})$), the (mathematically derived) cost term for each parameter type in (\ref{biccost}) depends on its specific \textit{effective} sample size ($\log(\text{effective sample size})$), which is in fact different for the topic probability and word probability parameters -- this effective sample size is $L_d$ for a topic probability and $\bar{L}_j$ for a word probability. This dependence is consistent with the fact that each parameter contributes differently to model complexity and hence should be penalized differently.

PTM invokes a generalized Expectation Maximization (GEM) algorithm \cite{Dempster1977},\cite{Meng1997} to optimize BIC. Assuming for now that the number of topics, $M$, is fixed, the GEM algorithm consists of the following expectation (E-step) and generalized minimization steps (Generalized M-step), iterated until convergence:

\textit{E-step}: PTM treats the topic of origin for each word $i$ in every document $d$ as the hidden data $z_{id}$. The E-step computes the expectation of the complete data BIC with respect to the posterior distribution of the hidden data given the current estimate of $\Theta$ and $\mathcal{H}$:
\begin{displaymath}
P(z_{id}^{(j)}=1|w_{id};\Theta^{(t)},\mathcal{H}^{(t)})
=\frac{\theta_{jd}v_{jd}p(w_{id}|z^{(j)}_{id}=1,\beta,u)}{\sum_{l=1}^{M}{\theta_{ld}v_{ld}p(w_{id}|z^{(l)}_{id}=1,\beta,u)}}.
\end{displaymath}

\textit{Generalized M-step}: In the M-step, we alternately optimize parameters given fixed structure and structure given fixed parameters. Each update in this step is guaranteed to decrease BIC (\ref{bic}).

\noindent 1) Minimizing BIC with respect to $\Theta$ given fixed structure is equivalent to maximizing the log-likelihood function. There are closed-form updates for all parameters. Optimization with respect to topic proportions is achieved by:
\begin{displaymath}
\theta_{jd}=\frac{v_{jd}\sum_{i=1}^{L_d}{P(z_{id}^{(j)}=1|w_{id};\Theta^{(t)},\mathcal{H}^{(t)})}}{\sum_{l=1}^{M}{\sum_{i=1}^{L_d}{P(z_{id}^{(l)}=1|w_{id};\Theta^{(t)},\mathcal{H}^{(t)})v_{ld}}}},\forall j,d.
\label{thetaupdate}
\end{displaymath}

Topic-specific word probabilities, are updated by:
\begin{displaymath}
\beta_{jn}=\frac{x_{jn}u_{jn}}{\mu_j},
\label{pj}
\end{displaymath}
where 
\[
\mu_{j}\triangleq\frac{\bar{x}_j}{1-\sum_{n=1}^{N}{(1-u_{jn})\beta_{0n}}},\bar{x}_j\triangleq \sum_{n=1}^{N}{x_{jn}u_{jn}},\forall j,
\]
\begin{displaymath}
x_{jn}\triangleq \sum_{d=1}^{D}{\sum_{{i=1:{w_{id}=n}}}^{L_d}{P(z_{id}^{(j)}=1|w_{id};\Theta^{(t)},\mathcal{H}^{(t)})v_{jd}}},\forall j,n. \nonumber
\end{displaymath}

The shared model is estimated via global frequency counts at initialization and we choose to not update it during GEM iterations:
\begin{displaymath}
\beta_{0n}=\frac{\sum_{d=1}^{D}{\sum_{i=1:{w_{id}=n}}^{L_d}{1}}}{\sum_{d=1}^{D}{L_d}},~~\forall n=1,...,N.
\label{p0}
\end{displaymath}

\noindent 2) Minimizing BIC with respect to $(\{u\},\{v\})$ given fixed $\Theta$: In this step, we cycle over all switches and visit each switch one by one. At each step, we trial-change one switch, check the change in BIC $(\Delta BIC = \Delta Cost - \Delta L)$, and only accept that change if it decreases BIC. This process is repeated until no further changes occur. 

In addition to estimating the binary switches $(u,v)$ and model parameters $\Theta$, minimizing BIC provides a sensible, objective way to compare models with different number of topics and properly choose model order. To achieve this, \cite{Soleimani2014} proposed a top-down approach to search over possible model orders. PTM is first trained with $M_{max}$ topics and then the number of topics is decreased one by one, removing the topic that has the minimum overall mass at each step. Then, the model with number of topics $M^*$ achieving the minimum BIC value is chosen as the optimal model for the corpus. 

\textit{Inference on Test Documents:} After learning topics on the training set, we can estimate topic distributions for documents in a test corpus $\mathcal{D}_{t}$. In this case, the word probabilities and $u$ switches in the model are fixed at their estimated values from the training phase. The only parameters to estimate then are $\Theta_{t} = \{\theta_{jd}~\forall j, \forall d \in \mathcal{D}_{t}\}$ and $\mathcal{H}_{t} = \{v_{jd}~\forall j, \forall d \in \mathcal{D}_{t}\}$. This is done by minimizing the same BIC objective function following the algorithm described above for training PTM, except that in the M-step we only update $\Theta_{t}$ and $\mathcal{H}_{t}$. That is, for test documents, we minimize the
BIC objective function specialized to the case where there is parametric dependence only on $\Theta_{t}$ and $\mathcal{H}_{t}$.  In this case, the BIC objective can be shown to reduce to:  
\begin{flalign}
\label{testbic}
BIC(\mathcal{D}_{t},\mathcal{H}_{t},\Theta_{t}) &= \frac{1}{2}\sum_{d\in \mathcal{D}_{t}}^{}{(M_d-1)\log(\frac{L_d}{2\pi})}\\
&+\sum_{d\in \mathcal{D}_{t}}^{}{\log\binom{M}{M_d}} - \log(p(\mathcal{D}_{t}|\mathcal{H}_{t},{\Theta}_{t})).\nonumber
\end{flalign}

\section{Anomalous Topic Discovery}
\label{ATDsection}
In this section, we introduce our algorithm for detecting anomalous topics. We assume that we have a collection of normal documents which sufficiently characterizes all normal topics. We learn PTM as described in section \ref{PTMsection} on this training corpus to discover the normal topics. Then, in the detection phase, our goal is to detect any and all patterns in the test corpus which are anomalous (unusual) with respect to the normal topics.

In our proposed algorithm, we detect anomalous topics in the test set one by one. That is, at each step, we detect the cluster of test documents $\mathcal{S}$ (candidate anomalous cluster) that exhibits the pattern with maximum ``deviance" from normal topics. Then, we conduct a statistical test to measure the significance of $\mathcal{S}$ and the topic exhibited by it, compared to the \textit{normal} topics hypothesis. If the cluster candidate is determined to be significantly anomalous, we declare it as detected, we remove all documents in $\mathcal{S}$ from the test set, and then repeat this process until no statistically significant anomalous topic is found.

Supposing for now that a candidate anomalous cluster $\mathcal{S}$ is given, {\it under the null hypothesis}, we posit that every document in $\mathcal{S}$ is well characterized solely using the normal topics. Accordingly, we determine $\Theta_{0} = \{\theta_{jd}~\forall j=1,..,M, \forall d \in \mathcal{S}\}$ and $\mathcal{H}_{0} = \{v_{jd}~\forall j=1,...,M, \forall d \in \mathcal{S}\}$ by minimizing with respect to the objective function (\ref{testbic}), where in this case $\mathcal{D}_{t} = \mathcal{S}$.  $\mathcal{M}_0 = \{\Theta_0,\mathcal{H}_0\}$ constitutes our null model. A normal document, is expected to have high likelihood under the null model. A document with anomalous contents, however, will have low likelihood under $\mathcal{M}_0$. Thus, as a quantitative measure to characterize how well documents in $\mathcal{S}$ fit under the null model, we compute $l_0(\mathcal{S}) =  \sum_{d \in \mathcal{S}}\log(p(d|\mathcal{M}_0)) = \sum_{d \in \mathcal{S}}l_0(d)$. 

Our {\it alternative hypothesis} is that $\mathcal{S}$ contains one new topic which is significantly different from the $M$ normal topics in $\mathcal{M}_0$. Therefore, to capture this unusual topic, we build our alternative model $\mathcal{M}_1$ by adding one topic to the null model. The likelihood of the candidate cluster $\mathcal{S}$ under $\mathcal{M}_1$ is:
\begin{flalign}
p(\mathcal{S}|\mathcal{M}_1) = \prod_{d\in \mathcal{S}}\prod_{i=1}^{N_d}\bigg[&\sum_{j=1}^{M}v_{jd}\theta_{jd}p(w_{id}|z^{(j)}_{id}=1,\beta,u)\nonumber\\
&+\theta_{(M+1)d}\hat{\beta}^{\hat{u}_{w_{id}}}_{w_{id}}\beta^{1-\hat{u}_{w_{id}}}_{0w_{id}}\bigg],
\label{alt_lkh}
\end{flalign}
where $\sum_{n=1}^{N}\hat{\beta}^{\hat{u}_n}_{n}\beta^{1-\hat{u}_n}_{0n} = 1$, $\sum_{j=1}^{M}v_{jd}\theta_{jd}+\theta_{(M+1)d}=1~\forall d \in \mathcal{S}$, and where, for concision, we have denoted $\beta_{(M+1)n}$ and $u_{(M+1)n}$ by $\hat{\beta}_{n}$ and $\hat{u}_n$, respectively.

The parameters of the first $M$ topics $\{\beta_{jn}, u_{jn},~\forall j=1,...,M, \forall n\}$ as well as the shared model $\{\beta_{0n} \forall n\}$ in (\ref{alt_lkh}) are the same as those in the null model, as estimated on the training set. That is, our alternative model is a PTM with $M+1$ topics -- $M$ normal topics and one candidate anomalous topic. We choose this specific structure for the alternative model consistent with the assumption that anomalous documents need not only contain anomalous contents -- only a subset of an anomalous document may contain novel topics, with the remaining words well-generated from normal topics. Note that in the generative process of PTM a different topic can be used to generate each word in a document. Thus, the first $M$ topics can be used to model normal contents of the documents (words that appear under ``usual'' contexts) while the new topic (topic $M+1$) captures the anomalous content. 

Parameters of the alternative model $\mathcal{M}_1$ that require estimation on $\mathcal{S}$ are $\Theta_1= \{\hat{\beta}_{n}\forall n, \theta_{jd}~j=1,...,M+1, d \in \mathcal{S}\}$ and $\mathcal{H}_1 = \{\hat{u}_n \forall n, v_{jd}~j=1,...,M, d \in \mathcal{S}\}$. Since topic $M+1$ is the (candidate) anomalous topic exhibited by $\mathcal{S}$, 
under the alternative hypothesis it should be active in all documents in $\mathcal{S}$. Thus, we fix $v_{(M+1)d} = 1~\forall d \in \mathcal{S}$. That is, under the alternative hypothesis, we assume that \textit{each} document in $\mathcal{S}$ contains words that are explained by the new topic (otherwise, there is no reason for inclusion of a document in $\mathcal{S}$). 

For a given candidate cluster $\mathcal{S}$, estimating $\Theta_1$ and $\mathcal{H}_1$ is achieved by minimizing (\ref{bic}) (for the choice $\mathcal{D} = \mathcal{S}$) using the algorithm given for training PTM, but wherein parameters of the first $M$ topics are kept fixed and we only update parameters of the alternative model. For this case, the BIC cost function specializes to:
\begin{flalign}
BIC(\mathcal{S}&,\Theta_1,\mathcal{H}_1)=\sum_{d\in \mathcal{S}}^{}{\log\binom{M+1}{M_d}}+NH(\frac{N_{(M+1)}}{N})\log(2)\nonumber\\
&+\frac{1}{2}\sum_{d\in \mathcal{S}}^{}{(M_d-1)\log(\frac{L_d}{2\pi})}+\frac{1}{2}{N_{(M+1)}\log(\frac{\bar{L}_{(M+1)}}{2\pi})}\nonumber\\
& - \log(p(\mathcal{S}|\Theta_1,\mathcal{H}_1)),
\label{bicM1}
\end{flalign}
where $N_{(M+1)}$ is the number of topic-specific words in the new topic and $\bar{L}_{(M+1)}=\sum_{d\in\mathcal{S}}L_d$ is the total length of documents in $\mathcal{S}$.

We measure goodness of fit of $\mathcal{M}_1$ on $\mathcal{S}$ by computing $l_1(\mathcal{S}) = \log(p(\mathcal{S}|\mathcal{M}_1)) = \sum_{d \in \mathcal{S}}\log(p(d|\mathcal{M}_1)) = \sum_{d \in \mathcal{S}}l_1(d)$. We evaluate the anomaly score for cluster $\mathcal{S}$ via $\text{score}(\mathcal{S}) = l_1(\mathcal{S})-l_0(\mathcal{S}) = \sum_{d \in \mathcal{S}}(l_1(d)-l_0(d))$. This anomaly score, in fact, measures the degree of deviance of the new topic from the normal topics, exhibited on $\mathcal{S}$. Using this score function, we can test the normal hypothesis $H_0:~\mathcal{S}\sim \mathcal{M}_0$ versus the alternative hypothesis $H_1:~\mathcal{S}\sim \mathcal{M}_1$. 

In practice, however, $\mathcal{S}$ is unknown and has to be discovered by searching over the test documents. Since the size of $\mathcal{S}$ is not known, a na\"{\i}ve search would involve choosing the cluster with highest score over all $2^{|\mathcal{D}_t|}$ subsets of the test set. This is clearly practically infeasible for any sizeable test corpus. Instead, in this paper, we propose an iterative algorithm to jointly search for the most anomalous cluster $\mathcal{S}$ in a greedy fashion and learn the parameters of $\mathcal{M}_1$ in a computationally tractable way.

We begin constructing $\mathcal{S}$ by choosing a document in the test set which has the lowest likelihood under the null model. We normalize for the length of documents and choose document $d^* = \argmin_{d\in \mathcal{D}_t} \frac{1}{L_d}l_0(d)$ as the first document in $\mathcal{S}$. We then learn parameters of the alternative model $\mathcal{M}_1$ on this document. 

Then, we continue our alternating algorithm by searching for the next best document to add to the cluster and then re-optimizing the alternative model. At each step, to choose the next document to include in $\mathcal{S}$, we first compute the log-likelihood of each of the remaining test documents under the current alternative model and compute the relative change in log-likelihood under the null and alternative models; {\it i.e.} $\Delta l(d) = \frac{l_1(d)-l_0(d)}{|l_0(d)|}~\forall d \in \mathcal{D}_t - \mathcal{S}$. We choose the document $d^*$ with highest $\Delta l$ as a {\it candidate} document ($d^* = \argmax_{d}\Delta l(d)$) to add to the cluster. We then perform a statistical test (Algorithm \ref{BSSingleDoc} in the sequel) and only add $d^*$ to $\mathcal{S}$ if our test determines that $d^*$ significantly belongs to $\mathcal{S}$. If the test reveals that contents of $d^*$ are not significantly related to the anomalous topic, we do not add $d^*$ to $\mathcal{S}$ and we stop adding further documents to the cluster.

At each step, after adding a new document to $\mathcal{S}$, we re-initialize all parameters of the alternative model on $\mathcal{S}$. We initialize word probabilities ($\hat{\beta}$) of the new topic via frequency counts and initialize $\hat{u}_n$ to 1 for all words that occur in documents in $\mathcal{S}$. We also initialize topic proportions in all documents in $\mathcal{S}$ consistent with topic $M+1$ being the dominant topic (proportions for other topics are all set to a small value, close to zero). We then train the alternative model $\mathcal{M}_1$ by minimizing (\ref{bicM1}) and then compute the anomaly score of the cluster $\text{score}(\mathcal{S}) = \sum_{d\in \mathcal{S}}(l_1(d)-l_0(d))$. 

After growing of $\mathcal{S}$ has terminated, we conduct another statistical test (Algorithm \ref{BSSigTest} in the sequel) to measure the significance of the anomaly score $\text{score}(\mathcal{S}) = \sum_{d\in \mathcal{S}}(l_1(d)-l_0(d))$. If $\mathcal{S}$ is found significantly anomalous, the cluster is reported as detected and we then remove all documents in $\mathcal{S}$ from the test set; the algorithm is then repeated on the new test set, until no significant cluster is found. Algorithm \ref{algvb} summarizes our ATD method. 

Topic $M+1$ in the alternative model is thus the anomalous pattern that prevails in documents in $\mathcal{S}$ and the set of topic-specific words under that topic are the subset of salient features in the cluster. 

\setlength{\textfloatsep}{1mm}
\begin{algorithm}[t]
\small
\caption{ATD: Anomalous Topic Discovery}
\label{algvb}
\begin{algorithmic}[1]
\STATE Input: Test set $\mathcal{D}_t$ and PTM with $M$ normal topics
\STATE Learn $\mathcal{M}_0=\{\Theta_0,\mathcal{H}_0\}$ on $\mathcal{D}_t$ 
\STATE Compute $l_0(d)~\forall d\in \mathcal{D}_t$
\REPEAT
\STATE Set $\mathcal{S} = \emptyset$
\STATE Choose $d^* = \argmin_{d\in \mathcal{D}_t} \frac{1}{L_d}l_0(d)$
\REPEAT
\STATE Set $\mathcal{S}\leftarrow \mathcal{S}\cup\{d^*\}$
\STATE Learn $\mathcal{M}_1=\{\Theta_1,\mathcal{H}_1\}$ on $\mathcal{S}$
\STATE Compute $l_1(d)~\forall d\in \mathcal{D}_t-\mathcal{S}$
\STATE Choose $d^* = \argmax_d \frac{l_1(d)-l_0(d)}{|l_0(d)|}$
\STATE Test significance of topic $M+1$ in $d^*$ (Algorithm \ref{BSSingleDoc})
\UNTIL{Topic $M+1$ is insignificant in $d^*$}
\STATE Compute $\text{score}(\mathcal{S})$
\STATE Test significance of $\mathcal{S}$ (Algorithm \ref{BSSigTest})
\STATE $\mathcal{D}_t\leftarrow \mathcal{D}_t-\mathcal{S}$
\UNTIL{$\mathcal{S}$ is insignificant}
\STATE Output: Discovered cluster $\mathcal{S}$ with significance measure p-value($\mathcal{S}$).
\end{algorithmic}
\end{algorithm}
\normalsize

\subsection{Determining Significance of the New Topic in a Candidate Document}
\label{BSStopsection}
At each step of our algorithm to construct the cluster $\mathcal{S}$, we need to determine if a candidate document $d^*$ significantly belongs to $\mathcal{S}$. In this section, we describe our algorithm to detect such significant documents and determine when to stop adding documents to a putative anomalous cluster. 

There are different possible methods to determine if a document significantly belongs to $\mathcal{S}$. One na\"{\i}ve approach is to consider each document a random draw from a multinomial distribution over all words in the dictionary and then use Pearson's chi-squared test to determine significance of the difference between the observed counts (words in document $d^*$) and the expected counts (word probabilities under the null or alternative models) \cite{Pearson1900},\cite{Agresti1996}.  A major problem with this approach is that the length of each document $L_d$ is typically much smaller than the vocabulary size $N$. While Pearson's chi-squared test relies on $L_d \rightarrow \infty$, in our problem generally $L_d \ll N$. 

In this paper, we follow a more practical approach by proposing a bootstrap algorithm. First, we note that since the major difference between the null and alternative models is the new topic, our decision on whether to include the candidate document in the cluster or not can be reliably made based on the contribution of the new topic in modeling words in the candidate document. That is, if the new topic is not used in modeling a significant percentage of the words in the document, it is sufficient to rely on the null model to describe all contents of this document. 

To determine the contribution of the new topic, we use the expected values of the latent variables in our EM algorithm. Note that expectation of the latent variables gives the posterior probability that topic $j$ generates the $i$th word in document $d$, $P(z^{(j)}_{id}=1|w_{id},\Theta,\mathcal{H})$. Using these probabilities, we hard-assign each word in $d^*$ to the topic that has the highest posterior and then count the number of words in the candidate document which are assigned to topic $M+1$; i.e. we compute the empirical proportion of the new topic $\hat{\theta}_{d^*} = \frac{1}{L_{d^*}}\#\{i: P(z^{(M+1)}_{id^*}=1|w_{id^*},\Theta,\mathcal{H})>P(z^{(j)}_{id^*}=1|w_{id^*},\Theta,\mathcal{H})\forall j =1,...,M\}$. If this number is very small, it indicates that the new topic does not play an important role in this document and suggests that the document should not be included in $\mathcal{S}$. 

Thus, in this step, we test the null hypothesis that topic $M+1$ is insignificant in document $d^*$ versus the alternative hypothesis that it is significant. To test this hypothesis, we conduct a bootstrap algorithm to generate a set of \textit{normal} documents from the null model, compute the topic contribution in those documents, and compare them with the topic contribution in the candidate document.
\setlength{\textfloatsep}{3mm}
\begin{algorithm}[t]
\footnotesize
\caption{Generating a Bootstrap Document}
\label{bsdocgen}
\begin{algorithmic}[1]
\STATE Input: Document $d^*$, validation corpus $\mathcal{D}_v$, topic proportions $\theta_{0d^*}$ and $\theta_{0d}~\forall d\in \mathcal{D}_v$ under the null model
\STATE Compute $\rho_{d^*}(d)=\frac{\underline{\theta}^T_{0d}\underline{\theta}_{0d^*}}{\|\underline{\theta}_{0d}\|\|\underline{\theta}_{0d^*}\|}~\forall d\in \mathcal{D}_v$
\STATE Set $\mathcal{D}' = \{d'': d''=\argmax_d \rho_{d^*}(d)\forall d \in \mathcal{D}_v \}$
\STATE Choose a document $d' \sim \text{uniform}(\mathcal{D}')$
\STATE Sample $w_{ib}\sim \text{Uniform}(w_{1d'},...,w_{L_{d'}d'})~\forall i=1,...,L_{d^*}$.
\STATE Output: Document $d_b = \{w_{1b},...,w_{L_{d^*}b}\}$
\end{algorithmic}
\end{algorithm}

There are two basic bootstrap approaches for generating documents from the null model -- parametric and non-parametric approaches. In the parametric approach, we can generate a document by following the generative process for PTM described in section 2. However, our experiments show that this approach results in unrealistic documents. In generating a bootstrap document in this approach, we need to randomly draw $L_d$ word samples from the N-dimensional multinomial distribution of topics. But, since $L_d \ll N$, it is very likely that documents generated by this method are collections of unrelated words that do not resemble any normal document. Instead, in this paper we follow a \textit{non-parametric} approach to generating bootstrap documents.

We first note that the training set only includes normal documents. Therefore, we could use the training set as a pool to generate bootstrap documents. Alternatively, we can hold out a portion of the training set for the purpose of generating bootstrap samples -- since the training set is used in learning the null model, also using it in our bootstrap algorithm may introduce bias in our significance test. In this paper, we take the latter approach and separate some documents from the training set, keeping them as a validation set for generating bootstrap documents. 

To conduct a fair bootstrap test, we need to ensure that the bootstrap documents have similar topic proportions to those of the candidate document under the null model. Moreover, the bootstrap documents and the candidate document should have the same length. To ensure that both these conditions are satisfied we propose the following non-parametric approach to generate bootstrap documents. First, from the set of validation documents we choose a document that has the highest similarity to the candidate document based on their topic proportions under the null model. We compute similarity between topic proportions of a validation document $d$ and the candidate document $d^*$ using the Cosine similarity measure: $\rho_{d^*}(d)=\frac{\underline{\theta}^T_{0d}\underline{\theta}_{0d^*}}{\|\underline{\theta}_{0d}\|\|\underline{\theta}_{0d^*}\|}$ where $\underline{\theta}_{0d}$ and $\underline{\theta}_{0d^*}$  are $M$ dimensional vectors of topic proportions under the null model, {\it i.e.} $\underline{\theta}_{0d} = (\theta_{1d},\theta_{2d},\ldots,\theta_{Md})^T$. We find the document $d' = \argmax_{d} \rho_{d^*}(d)~\forall d=1,..,D_v$ where $D_v$ is the number of documents in the validation set. Note that due to sparsity of topic proportions, the validation document with maximum similarity to the candidate document may not be unique. In this case, we define $\mathcal{D}' = \{d'': d''=\argmax_d \rho_{d^*}(d)\forall d \in \mathcal{D}_v \}$ and randomly choose one of the documents from $\mathcal{D}'$, $d'\sim \text{uniform}(\mathcal{D}')$.  Then, from the $L_{d'}$ words in document $d'$, we randomly choose $L_{d^*}$ words with replacement. Algorithm \ref{bsdocgen} summarizes the procedure used to generate a bootstrap document for a given document $d^*$.
\begin{algorithm}[t]
\setlength{\textfloatsep}{3mm}
\footnotesize
\caption{Testing Significance of Topic $M+1$ in Document $d^*$}
\label{BSSingleDoc}
\begin{algorithmic}[1]
\STATE Input: Candidate document $d^*$, validation corpus $\mathcal{D}_v$, alternative model $\mathcal{M}_1$, and null model $\mathcal{M}_0$
\STATE Compute the empirical proportion of the new topic $\hat{\theta}_{d^*}$
\FOR{$b=1$ to $B_1$}
\STATE Generate bootstrap document $b$ (Algorithm \ref{bsdocgen})
\STATE Learn topic proportions $\theta_b$ under $\mathcal{M}_1$.
\STATE Compute $\hat{\theta}_{b}$
\ENDFOR
\STATE Output: $t(\hat{\theta}_{d^*}) = \frac{\# \{b: \hat{\theta}_{b} < \hat{\theta}_{d^*}\}+1}{B_1+1}$
\end{algorithmic}
\end{algorithm}

After generating a bootstrap document $b$, we learn the topic proportions of that document under the alternative model and determine the empirical proportion of the new topic $\hat{\theta}_{b}$. We then repeat this process $B_1$ times and compute a statistic $t(\hat{\theta}_{d^*})$ to measure significance of the new topic in document $d^*$; $t(\hat{\theta}_{d^*}) = \frac{\# \{b: \hat{\theta}_{b} < \hat{\theta}_{d^*}\}+1}{B_1+1}$. We stop adding documents to the cluster $\mathcal{S}$ if $t(\hat{\theta}_{d^*})$ is less than a threshold value. 

Experimentally, we find that a candidate document $d^*$ is found insignificant only when $\hat{\theta}_{d^*}$ is small (e.g. less than $0.2$). Thus, to reduce computational burden, we only perform this bootstrap test if $\hat{\theta}_{d^*}$ $<$ 0.2. Also, to have a more robust termination criterion, we terminate further growing of $\mathcal{S}$ if $t(\hat{\theta}_{d^*})$ is smaller than the pre-set threshold over two consecutive steps. 

At the very first few steps of constructing cluster $\mathcal{S}$, the estimate of the parameters of the new topic ($\hat{\beta}$) could be very unreliable since they are estimated based on only a few documents. Thus, in general, it is possible that our test to determine significance of $\hat{\beta}$ in a new candidate document could terminate construction of $\mathcal{S}$ prematurely. Although Algorithm \ref{BSSigTest} (discussed in the sequel) can still in principle reject such small candidate clusters as significantly anomalous, to avoid false detections, we perform the test to terminate further growing of $\mathcal{S}$ only when $\mathcal{S}$ has at least 4 documents. Algorithm \ref{BSSingleDoc} summarizes our bootstrap test for detecting significance of the new topic in candidate document $d^*$.

\subsection{Significance Test for a Cluster}
After growing of a cluster has terminated, we need to determine whether the anomalous topic exhibited by the documents in that cluster is significant. Again, we note that due to small sample size, asymptotic distributions commonly known for the likelihood ratio test \cite{Wilks1938} do not hold. Instead, we perform bootstrap testing to compare significance of a candidate cluster $\mathcal{S}$ to normal clusters.

We use an algorithm similar to the procedure described in section \ref{BSStopsection} for generating bootstrap documents. We generate $|\mathcal{S}|$ bootstrap documents ($\mathcal{S}_b$) based on the null distribution from a collection of validation documents and compare the likelihood ratio score of this bootstrap cluster with that of the candidate cluster. Similar to the last section, for each document in the candidate cluster $\mathcal{S}$, we generate a bootstrap document with similar topic proportions under the null model and with the same length. Then, we learn the alternative model ($u$ switches and topic-specific word probabilities of the new topic as well as topic proportions under the alternative model) and compute the log-likelihood ratio score $score(\mathcal{S}_b)$. We repeat this process $B_2$ times and compute the empirical p-value to measure significance of the candidate cluster; $\text{p-value}(\mathcal{S}) = \frac{\# \{b: \text{score}(\mathcal{S}_b) > \text{score}(\mathcal{S})\}+1}{B_2+1}$. Algorithm \ref{BSSigTest} summarizes the procedure we use in this section.
\begin{algorithm}[t]
\setlength{\textfloatsep}{3mm}
\footnotesize
\caption{Testing Significance of $\mathcal{S}$}
\label{BSSigTest}
\begin{algorithmic}[1]
\STATE Input: candidate cluster $S$, score($\mathcal{S}$)
\FOR{$b=1$ to $B_2$}
\STATE Set $\mathcal{S}_b = \emptyset$
\FOR{$d=1$ to $|\mathcal{S}|$}
\STATE Generate a bootstrap document $d_b$ for document $d$ (Algorithm \ref{bsdocgen})
\STATE $\mathcal{S}_b\leftarrow \mathcal{S}_b \cup \{d_b\}$
\ENDFOR
\STATE Learn $\mathcal{M}_0$ and $\mathcal{M}_1$ on $\mathcal{S}_b$
\STATE Compute score($\mathcal{S}_b$)
\ENDFOR
\STATE Output: $\text{p-value}(\mathcal{S}) = \frac{\# \{b: \text{score}(\mathcal{S}_b) > \text{score}(\mathcal{S})\}+1}{B_2+1}$
\end{algorithmic}
\end{algorithm}

\section{Experimental Results}
\label{Expsection}
In this section, we compare performance of our algorithm against four baseline methods on a synthetic data set and two text corpora. We use ground-truth class labels of the data sets to define anomalous classes. In each data set, we choose some classes as anomalous and take all documents from those classes out of the training and validation sets. We then randomly select some documents from normal classes (excluding documents chosen for the training set) and some documents from anomalous classes to create the test set. Our goal is to detect clusters of documents from the anomalous classes in the test set. Note that we only use the class labels for creating the training and test sets and for evaluating test set performance.

For group AD performance, we measure recall and precision for each cluster $\mathcal{S}$.  These are, respectively, the number of true detected anomalies from the majority anomalous class in that cluster divided by the total number of true anomalies (from the majority anomalous class in the cluster) and the number of true detected ones (again from the majority anomalous class in the cluster) divided by the size of the cluster. We also compute area under the recall/precision curve (AUC) as well as the F1-measure, which is equal to $2\times\text{recall}\times\text{precision}/(\text{recall}+\text{precision})$.

\subsection{Baseline Methods}
\textit{Group (Cluster) Anomaly Detection:} We compare our method for group anomaly detection against four baseline methods. Unlike our algorithm, none of the baseline methods have a statistical test to determine significance of a candidate cluster; they only provide an anomaly score for every cluster. To have a fair comparison with ATD, for each method, we sort the detected clusters based on their anomaly score and take as anomalous the same number of top clusters as detected by our method. Then, we compute the AUC and F1 measures for those clusters declared as anomalous. We also determine hyper-parameters of the four baseline methods by performing a grid search and picking the values that achieve the best average F1-measure on the top detected clusters for the \textit{test} set. This essentially gives ``upper bound'' performance for each baseline method of comparison.

\textit{Individual Point Anomaly Detection:} In addition to group AD, we also compare our method in detecting individual anomalies against some of our baseline methods, which nominally are individual point AD techniques. For our method, we take individual documents in all detected anomalous clusters in the order in which they are added to their clusters and compute the AUC and F1 measures. That is, we take $N_0 = \sum_{i}|\mathcal{S}_i|$ test documents as individual point anomalies by our method, where $|\mathcal{S}_i|~\forall i$ is the size of the $i$-th anomalous cluster detected by ATD. To have a fair comparison, for each baseline method, we sort all test samples based on their anomaly scores and take the same number of points ($N_0$) with highest anomaly scores as anomalous. We then compute AUC and F1 measures based on these selected samples. An effective individual point AD method should have many true anomalies and few false detections in its top $N_0$ points. We determine hyper-parameters of each baseline method by performing a grid search and choosing the values which achieve the highest F1-measure.

\subsubsection{Mixtures of Multinomial Mixture Models (M4)}
Following the idea proposed in \cite{Xiong2011},\cite{Xiong2011a}, we use Mixtures of Multinomial Mixture Models (M4) as one baseline method. Here, we assume that there are $G$ groups and $T$ typical genres in the data set. The generative process for this model is as follows:
\begin{enumerate}[leftmargin=*]
\item For each group $g=1,\ldots,G$
\item Draw a genre $Y_g \sim \text{Multinomial}(\pi), ~Y_g\in\{1,...,T\}$.
\item For each document $d_g = 1,...,D_g$ in this group:
\begin{enumerate}
\item Draw a topic distribution according to the genre $y_g$: \\$\theta_{d_g} \sim \text{Dirichlet}(\alpha_{y_g1},...,\alpha_{y_gM})$.
\item For each word $i=1,\ldots,L_d$
\begin{enumerate}
\item Select a topic $j \sim \text{Multinomial}(\theta_{1{d_g}}...,\theta_{M{d_g}})$.
\item Choose a word $w_{i{d_g}}\sim\text{Multinomial}(\beta_{j1},...,\beta_{jN})$.
\end{enumerate}
\end{enumerate}
\end{enumerate}

Note that the prior probability on topic proportions of the documents in each group depends on the genre selected for that group. Also, we treat word probabilities as deterministic parameters to be estimated rather than random variables. Latent variables in this model include the genre for each group $Y_g$, topic proportions for each document $\theta$, and the topic of origin for each word in every document. We integrate out these latent variables and estimate the word probabilities $\beta$, genre proportions $\pi$, and the Dirichlet parameters for each genre $\alpha$. Similar to \cite{Xiong2011}, exact inference in this model is intractable and, instead, we use mean-field variational inference \cite{Jordan1999}. 

Note that this method (similar to \cite{Xiong2011}) does not learn group memberships of the documents. Instead, we need to cluster the documents into $G$ groups before applying this method. Therefore, there are three hyper-parameters ($T$, $G$, $M$) that need to be determined. However, for simplicity, we assume $G=M$. To do the clustering, we learn an LDA model with $M$ topics on the training data and hard-assign each document to the topic that has the highest proportion in that document. This way, we can find $M$ ($G$) clusters on the training data. After learning the clusters, we learn the M4 model parameters on the training set. 

In the detection phase, similar to the training step, we first perform clustering into $G$ groups and then compute an anomaly score for each group. The anomaly score is similar to the one suggested in \cite{Xiong2011}. 
The score for document $d$ is $\text{score}(d) = -E_{p(\theta_d|\cdot)}\big[\log p(\theta_d|\cdot)\big],$ where $p(\theta_d|\cdot)$ is the posterior topic proportions of document $d$. Since the exact posterior is not available, we approximate this score using the variational distribution of $\theta_d$. The score of a cluster is then the average of the scores of all documents in that cluster.

\subsubsection{Likelihood-Based Method (LB)}
One na\"{\i}ve way of detecting anomalies is by sorting the test documents based on their likelihood under the ATD null model normalized by their length. Naturally, we expect the documents with lowest normalized likelihoods to be the most anomalous ones. Similar to M4, we use LDA to perform clustering on the test data. In this method, the anomaly score for each cluster is the average of the normalized log-likelihood (under the ATD null model) of all documents in that cluster. The number of LDA topics (number of clusters) is the sole hyper-parameter of this method. We compare our method against LB for both group AD and individual point AD.

\subsubsection{One Class Support Vector Machines}
A widely used method for individual point anomaly detection is one-class support vector machines \cite{Scholkopf2001},\cite{Manevitz2001}. In this paper, we use one-class SVM models with linear and RBF kernels. We also represent each document $d$ by its normalized bag of words features; i.e an $N$-dimensional vector $(x_{d1},...,x_{dN})$ where $x_{dn}$ is the count of word $n$ in document $d$ normalized by the length of the document. We cluster test documents into $M$ groups using LDA. As the anomaly score of each cluster, we count the number of individual document anomalies detected by one-class SVM in that cluster. The hyper-parameters of the SVM (one-class SVM hyper-parameter and the hyper-parameter of the RBF kernel) and $M$ are the hyper-parameters of this method. We report the performance of this method for group AD and individual point AD.

\subsubsection{Nearest Neighbor Method (NN)}
\cite{Zhao2009} proposes a method for anomaly detection based on nearest neighbor graphs. In this method, we first compute the distance between each pair of training samples and denote the distance between every data point $d$ to its $K$-th nearest neighbor by $R_K(d)$. Similarly, for each test sample $d_t$, we compute $R_K(d_t)$; i.e. the $K$-th smallest distance between $d_t$ and all the training points. Then, we compute a p-value for each test sample $d_t$ by comparing $R_{K}(d_t)$ to $R_K(d)$ for all data points $d=1,...,D$ in the training set:
\[
\text{p-value}(d_t) = \frac{\#\{d:R_K(d_t)<R_K(d)\}}{D}.
\]

Here, we experiment with two different document representations: 1) In NN (BOW), we represent each document by its normalized bag of words (BOW) features. 2) In NN (LDA), we represent each document by its topic proportions under an LDA model learned on the training set. In both cases, we compute the distance between each pair of documents as one minus the Cosine similarity between their feature representations. We also use an LDA model to perform clustering on the test set. We compute the p-value of a candidate cluster as the average of p-values of all individual documents in that cluster. A group with smaller p-values has higher degree of anomaly. The number of nearest neighbors, $K$, that are used in computing the p-value and the number of topics $M$ are the two hyper-parameters of this model. We use NN as a baseline to compare with our method for both group AD and individual point AD.

\subsection{Synthetic Data}
We generated synthetic documents based on 10 normal topics on a dictionary with 3000 unique words. Under each topic, we chose 30 words as salient, with higher probabilities than other words. As the training data, we generated 3000 documents based on the generative process of LDA. For each training document, we chose one topic to be dominant with proportion equal to 0.85, with the rest of the topics having equal proportions. Each topic is dominant in 300 documents. Similarly, we generated 3000 documents for the validation set to generate bootstrap documents. Additionally, we generated two anomalous topics in a similar way. Our test set included 2060 documents, 200 documents generated from each normal topic and 30 documents from each anomalous topic. To make the detection problem more challenging, for the anomalous documents we chose two topics as the dominant ones -- one anomalous topic and one normal topic. That is, almost half of the words in each anomalous document were generated from the normal topics. 

We first learned the PTM null model on the training set. The optimal number of topics determined by minimizing BIC was 10, matching the true number of topics. We then ran our algorithm to detect clusters in the test set, one by one.  Table \ref{synAPD} shows the results. The first two clusters detected by ATD contain all documents from anomalous topics 12 and 11, respectively. The bootstrap algorithm shows that these two clusters are indeed anomalous compared to normal clusters. The p-value of the third cluster, however, is $0.34$ -- this cluster is not anomalous.  
\begin{table}[t]
\begin{center}
\caption{Results of ATD on Synthetic Data}
\label{synAPD}
\scriptsize
\begin{tabular}{c|c|c|c|c|c|c|c}
\hline \hline
\multirow{2}{*}{index} & \multirow{2}{*}{$|\mathcal{S}|$} & \multirow{2}{*}{label}	& \multirow{2}{*}{Recall} & \multirow{2}{*}{Precision} & \multirow{2}{*}{p-value} & \multicolumn{2}{ c }{\# salient words}	\\\cline{7-8}
 & 	&  &  &  &  & occurring & total \\\hline
1	&	32	&	12	&	1.0	&	0.94		&	$<$0.001	&	52	& 113 \\\hline
2	&	32	&	11	&	1.0	&	0.94		&	$<$0.001	&	64 	& 111 \\\hline
3	&	11	&	-	&	-	&	-		&	0.34	&	1 	& 2 \\
\hline \hline
\end{tabular}
\end{center}
\end{table}

\begin{table*}[t]
\begin{center}
\caption{Performance Comparison}
\label{synComp}
\scriptsize
\setlength\tabcolsep{6pt}
\begin{tabular}{|>{\centering\arraybackslash}p{6.5mm}|>{\centering\arraybackslash}p{8mm}|>{\centering\arraybackslash}p{8mm}|>{\centering\arraybackslash}p{8mm}|>{\centering\arraybackslash}p{8mm}|>{\centering\arraybackslash}p{8mm}|>{\centering\arraybackslash}p{8mm}|>{\centering\arraybackslash}p{8mm}||>{\centering\arraybackslash}p{8mm}|>{\centering\arraybackslash}p{8mm}|>{\centering\arraybackslash}p{8mm}|>{\centering\arraybackslash}p{8mm}|>{\centering\arraybackslash}p{8mm}|>{\centering\arraybackslash}p{8mm}|}
\hline 
~ & \multicolumn{7}{c||}{Group AD} & \multicolumn{6}{c|}{Individual AD}\\\hline
{\vspace{-0.5mm}Method} & \vspace{-0.5mm}\textbf{ATD} & \vspace{-0.5mm}M4 & \vspace{-0.5mm}LB & SVM (linear) & SVM (RBF) & NN (LDA) & NN (BOW)	&	\vspace{-0.5mm}\textbf{ATD} & \vspace{-0.5mm}LB & SVM (linear)& SVM (RBF) & NN (LDA) & NN (BOW) \\\hline
\multicolumn{14}{|c|}{Synthetic Data Set}\\\hline
F1	&	\textbf{0.97}	&	0.13		&	0.66		&	0.61		&	0.76		&	0.24		&	0.24		&	\textbf{0.97	}	&	0.95		&	0.90		&	0.5		&	0.90		&	0.84 \\\hline
AUC	&	\textbf{0.97	}	&	0.08		&	0.31		&	0.31		&	0.41		&	0.04		&	0.04		&	\textbf{0.96	}	&	0.96		&	0.90		&	0.32		&	0.84		&	0.83 \\\hline 
\multicolumn{14}{|c|}{20-Newsgroup}\\\hline
F1	&	\textbf{0.86	}	&	0.39		&	0.57		&	0.41		&	0.41		&	0.26		&	0.50		&	\textbf{0.86	}	&	0.32		&	0.21		&	0.22		&	0.34		&	0.44 \\\hline
AUC	&	\textbf{0.72	}	&	0.15		&	0.23		&	0.25		&	0.25		&	0.07		&	0.17		&	\textbf{0.77	}	&	0.13		&	0.06		&	0.04		&	0.06		&	0.24 \\\hline
\multicolumn{14}{|c|}{Reuters}\\\hline
F1	&	\textbf{0.84	}	&	0.42		&	0.41		&	0.55		&	0.44		&	0.29		&	0.54		&	\textbf{0.90	}	&	0.61		&	0.48		&	0.54		&	0.62		&	0.69 \\\hline
AUC	&	\textbf{0.79	}	&	0.14		&	0.16		&	0.22		&	0.39		&	0.10		&	0.19		&	\textbf{0.88	}	&	0.43		&	0.32		&	0.18		&	0.43		&	0.54 \\\hline 
\end{tabular}
\end{center}
\end{table*}

Table \ref{synAPD} also shows the number of topic-specific words (salient features) of the anomalous topic discovered for each cluster. Note that salient words need not necessarily {\it occur} in any document in $\mathcal{S}$. A word can be salient because it does {\it not} occur in any document in $\mathcal{S}$ even though it is very likely to co-occur with other words in $\mathcal{S}$ under normal topics. In this case, documents in $\mathcal{S}$ are anomalous due to \textit{absence} of topic-specific words. We separately report the total number of and the number of occurring topic-specific words under each topic. Under each anomalous topic, approximately 3\% of the unique words are topic-specific. Also, by comparing with the ground-truth topic distributions, we can see that all high-probability words under each novel topic are among the occurring topic-specific words under that topic.

To further measure how well ATD detects anomalous topics, we sorted words based on their probability under each detected anomalous topic in descending order and report the median rank of the thirty ground-truth high probability words of that topic. The median rank of significant words of the first and second clusters are, respectively, 14.5 and 14. This shows that the topics discovered by ATD can detect the salient features under each topic. 

We compared ATD against the four baseline group AD methods. Table \ref{synComp} shows the F1-measure and AUC, both
averaged over the top two clusters, for all methods. We can see that ATD significantly outperforms all other methods. Since only part of each document contains anomalous contents, methods that detect anomalies based on the entire feature space fare poorly in detecting anomalous documents in this experiment. Also, all the baseline methods perform separate clustering and anomaly detection, which gives another reason for their poor performance. 

Table \ref{synComp} also compares ATD against LB, SVM, and NN in detecting individual anomalies. We can see that ATD also outperforms the baseline methods in this comparison. Since the dictionary size in this data set is relatively small, and the high probability words under different topics are mostly disjoint, each of the documents with anomalous topics has high individual degree of atypicality. Thus, Individual AD methods have relatively good performance. This experiment was mostly designed to evaluate performance of our method for group AD and to evaluate the accuracy of the detection of salient words under anomalous topics. The next two experiments better demonstrate inefficiency of individual AD methods when the feature space is high-dimensional and the anomalous topic manifests on a low-dimensional subspace.

\begin{table*}[t]
\begin{center}
\caption{Sample words from detected anomalous topics}
\label{SampleWrds}
\footnotesize
\begin{tabular}{p{.05cm}|p{8cm}|p{8cm}}
\hline \hline
\multicolumn{1}{r|}{index} & ``Hockey'' topic (20-Newsgroup corpus)	& ``coffee'' topic (Reuters data set) \\\hline
\multicolumn{1}{c|}{1} &	game, team, hockey, play, win, pit, year, player, post, espn	 &	coffe$^*$, export, quota, produc, meet, price, brazil, year, intern$^*$, colombia 	\\\hline
\multicolumn{1}{c|}{2} &	window, graphic, car, error, buy, found, case, god, machin$^*$, appreci	&	japanes, mine, reserv, profit, barrel, growth, reagan, energi, soviet, japan	\\\hline
\multicolumn{1}{c|}{3} &	develop, ground, convert, relat, book, drug, screen, modem, port, moral	&	tariff, taiwan, liquid, baker, field, deposit, yen, inc, explor, canadian	\\\hline
\multicolumn{1}{c|}{4} &	don, think, time, good, look, new, way, well, right, now	&	market, last, report, expect, month, one, tonn, offici, new, end	\\
\hline \hline
\end{tabular}\\
\end{center}
\vspace{0mm}
1. high probability occurring salient words, 2. low probability occurring salient words, 3. low probability non-occurring salient words, 4. high probability occurring shared words. $^{(*)}$ Note that the words reported here are the stemmed forms of the real words in the documents; e.g. ``coffe'', ``intern'', ``machin'' are the stemmed forms of ``coffee'', ``international'', and ``machine'', respectively.
\normalsize
\end{table*}

\subsection{20-Newsgroup Corpus}
In this section, we report the results of our comparison on the 20-Newsgroup corpus\footnote{\url{http://qwone.com/~jason/20Newsgroups/20news-bydate.tar.gz}}. We removed stop words, applied Porter stemming \cite{Porter1980}, and removed too long and too short documents. We treated documents from classes ``rec.sport.hockey'' and ``talk.politics.mideast'' as anomalous. Note that there are documents with similar labels such as ``rec.sport.baseball'' and ``talk.politics.misc'', which are treated as normal. While we can select any class labels as anomalous, we chose these two topic classes because they better represent the type of anomaly which we expect to encounter in real applications. Specifically, each of these anomalous classes is closely related to some normal topics, and they may even share some keywords. But each of these anomalous classes is significantly different from the normal topics with respect to some low-dimensional (salient) subset of the word space. Thus, the problem well-matches the scenario of ``cluster anomaly detection on low-dimensional feature subspaces''. Nevertheless, the comparison results reported here for these two classes are typical of the results we have observed for other (arbitrarily chosen) anomalous classes.
  
We also removed all documents with label ``misc.forsale" since documents from this class contain a diverse set of words and do not sufficiently capture a coherent topic. The training and validation data sets have, respectively, 7973 and 1138 documents. The test set includes 2277 normal documents and 149 and 109 documents, respectively, from classes ``rec.sport.hockey'' and ``talk.politics.mideast''. The dictionary size in this data set is 33565.

The minimum of the BIC objective function on the training data is achieved for PTM with 17 (normal) topics. We then use ATD on the test set for detecting anomalous clusters. Table \ref{20NGAPD} shows that ATD can accurately detect anomalous topics in this corpus. The first two clusters largely contain, respectively, documents from class ``rec.sport.hockey'' and ``talk.politics.mideast''. The bootstrap algorithm finds these two classes statistically anomalous with p-value $<$ 0.001, while the third class has insignificant p-value. 

Table \ref{20NGAPD} also reports the number of salient features under each topic. Note that approximately 1\% of the unique words in the dictionary are salient under each topic. We report some sample topic-specific and shared words from topic ``rec.sport.hockey'' in Table \ref{SampleWrds}. The top 10 high probability occurring topic-specific words from this topic are all related to the topic ``hockey'' and are used more frequently under this topic than any other topic. We also report 10 occurring and non-occurring low probability topic-specific words. These are the words that appear with less frequency (occurring words) than expected or do not occur (non-occurring words) in documents in the detected cluster. These are the words that tend to co-occur with some of the words that do appear in the cluster under normal topics. Table \ref{SampleWrds} also reports 10 shared high probability words. These words appear relatively frequently under this topic but their frequency is more or less the same under different topics; therefore, they are not salient (topic-specific) under this topic.

Table \ref{synComp} reports the results of our comparison with baseline methods. For each baseline method, we took the two clusters with highest anomaly score as anomalous and reported the average F1-measure and AUC over these two
clusters. We can see that ATD outperforms all baseline methods. We also compare our method against some baseline methods in detecting individual anomalies. For each method, we took the 215 test samples with highest anomaly scores and compared the performance with the same number of samples detected in the two clusters by ATD. Our method also achieves better results in detecting individual anomalies. Note that SVM and LB, which are nominally individual point AD methods, perform better on this data set when used for group AD than for individual AD. This shows that some anomalous documents are not easily detectable by these methods when considered individually and can only be discovered when they are jointly considered in clusters, along with other similar documents.
Note also that the performance advantage of ATD for Individual AD is much greater on 20-Newsgroup than for
the synthetic data experiment.  Again, we expect this is due to the fact that a much smaller percentage of the 
dictionary is salient to the anomalous topics in 20-Newsgroup, compared to the percentage in the synthetic data experiment.  We also note that NN(BoW) outperforms NN(LDA) on the real data sets.  Also, SVM(RBF) outperforms SVM(Linear) on some of the data sets.

\textit{Using LDA instead of PTM within ATD:} To examine the importance of PTM's word sparsity for group anomaly detection, we repeated our experiment on the 20-Newsgroup data set with a variant of our ATD algorithm which uses LDA instead of PTM for both null and alternative modeling. In this variant of ATD with LDA as the base topic model, we do not search for topic-specific words under each topic -- every word under each topic has its own free parameter. The search process for constructing $\mathcal{S}$ continued for 500 steps, at which point we manually stopped the process. This cluster of size 500 had AUC $= 0.038$ and F1-measure $= 0.16$. This poor performance shows that searching for anomalous clusters defined on the full feature space degrades the detection accuracy. To accurately detect anomalous topics, it is imperative that the ATD algorithm search for anomalous clusters on a low-dimensional subspace of the full feature space.

\begin{table}
\begin{center}
\caption{Results of ATD on 20-Newsgroup Data Set}
\label{20NGAPD}
\scriptsize
\begin{tabular}{L{0.2cm}|c|c|c|C{0.8cm}|c|c|c}
\hline \hline
\multirow{2}{*}{\hspace{-0.2cm}\vspace{+0.1in}index} & \multirow{2}{*}{$|\mathcal{S}|$} & \multirow{2}{*}{label}	& \multirow{2}{*}{Recall} & \multirow{2}{*}{\hspace{-0.1cm}\vspace{+0.1in}Precision} & \multirow{2}{*}{p-value} & \multicolumn{2}{ c }{\# salient words}	\\\cline{7-8}
 & 	&  &  &  &  & occurring & total \\\hline
1	&	115	&	Hockey	&	0.76		&	0.98		&	$<$0.001	&	344 &	429 \\\hline
2	&	100	&	MiddleEast	&	0.82		&	0.90		&	$<$0.001	&	230	&	286  \\\hline
3	&	5	&	-			&	-		&	-		&	0.201	&	6	& 	7 \\
\hline \hline
\end{tabular}
\end{center}
\end{table}

\subsection{Reuters Corpus}
In this section, we report the results on the Reuters-21578 data set\footnote{\url{http://www.daviddlewis.com/resources/testcollections/reuters21578/}}. We applied the usual stop word removal and Porter stemming on this data set and also removed too long and too short documents. We treated the two classes with labels ``ship'' and ``coffee'' as anomalous and kept all documents with these labels out of the training set. Some documents in this data set have more than one class label. In these cases, a document is counted as truly anomalous if it contains either of the anomalous class labels. The detection problem for this data set is potentially more challenging than
for 20-Newsgroup as some parts of the anomalous documents are from normal topics. For instance, some documents from class ``ship'' also have labels ``crude'' or ``grain'' which are two other major topics in this data set. Our training and validation data sets have 2091 and 139 documents, respectively. The test set contains 650 normal documents and 56 and 85 documents, respectively, from classes ``ship'' and ``coffee''. There are 9469 unique words in the dictionary.

The minimum of the BIC objective function on the training data is achieved for PTM with 21 topics. Table \ref{R52APD} shows the results of ATD. Again, our method can detect anomalous classes with high accuracy. The first two clusters detected by ATD (topics ``coffee'' and ``ship'', respectively) are anomalous (p-value $<$ 0.003) but the third cluster is not significant. The number of salient features under each topic is also reported in Table \ref{R52APD}. Under each topic, approximately 2\% of the words are identified as salient. We also report some sample topic-specific (occurring and non-occurring) and shared words under topic  ``coffee'' in Table \ref{SampleWrds}.
\begin{table}[th]
\begin{center}
\caption{Results of ATD on Reuters Corpus}
\label{R52APD}
\scriptsize
\begin{tabular}{c|c|c|c|c|c|c|c}
\hline \hline
\multirow{2}{*}{index} & \multirow{2}{*}{$|\mathcal{S}|$} & \multirow{2}{*}{label}	& \multirow{2}{*}{Recall} & \multirow{2}{*}{Precision} & \multirow{2}{*}{p-value} & \multicolumn{2}{ c }{\# salient words}	\\\cline{7-8}
 & 	&  &  &  &  & occurring & total \\\hline
1	&	97	&	coffee	&	0.99		&	0.87		&	$<$0.001	&	166	&	193 \\\hline
2	&	49	&	ship		&	0.77		&	0.77		&	0.003	&	160	&	180 \\\hline
3	&	5	&	-		&	-		&	-		&	0.11		&	16	&	19 \\
\hline \hline
\end{tabular}
\vspace{-0.15in}
\end{center}
\end{table}

Table \ref{synComp} shows the results of the comparison between our method and other baseline methods for both group and individual AD. ATD again has better average F1-measure and AUC on this data set.

\section{Discussion}
\label{disc_section}
On all data sets, we see that ATD can accurately detect the anomalous topics and (as verified for the synthetic data set, and anecdotally from Table 3) their salient features. Our proposed statistical test can also determine significance of each detected cluster efficiently and with low false detection rate. 

Anomalous topics can be highly similar to normal topics with respect to most of the features, but may exhibit atypical patterns only on a small subset of the words. We saw that approximately 1-2\% of the words are salient under each anomalous topic. That is, most of the words are essentially uninformative features for detecting the anomalous clusters. Therefore, using the entire feature space for detecting anomalous topics should, in general, degrade the detection accuracy. That is in fact what we observed on all data sets; we see that several group AD methods which detect anomalies based on the entire feature space have poor detection performance. Our method, in contrast, jointly detects the salient subspace and the anomalous samples defining a cluster. In all the experiments, we saw that this approach led to high detection accuracy. This is mostly achieved because of the parsimony and parameter sharing framework of PTM. Using PTM, we are able to jointly determine the set of salient words under each topic and estimate their word probabilities. We saw that the topic-specific words of the alternative model match the salient words of the (ground truth) anomalous topics (both on the synthetic data set and, anecdotally, on the real data sets based on Table 3). To further examine the importance of PTM in our detection algorithm, we modified ATD and replaced PTM by LDA as the base model. Experiments showed that this variant of ATD had very poor performance because, unlike for PTM, topics in LDA are estimated on the entire feature space, with one free parameter for each word under every topic. 

We note that unlike several existing group AD methods, our proposed approach jointly discovers candidate anomalous clusters and measures their level of atypicality; existing group AD methods can only measure an anomaly score for clusters detected by a separate algorithm. This is one of the sources of their suboptimal detection performance. Another source is the particular clustering algorithm that is used. In this paper, we used LDA to detect clusters for the baseline group AD methods. This is a reasonable choice, as LDA is one of the most widely used methods for identifying topical content in documents. However, as we have noted, unlike PTM, LDA does not achieve sparsity in the word (feature) space, and it achieves much {\it less} sparsity in topic presence in documents than PTM.  We believe a second source of performance advantage of ATD over the baseline methods is its adapted use of PTM, which achieves these two types of sparsities in the clustering solutions.  Particularly, sparsity in the word space is important, as in the text domain, topics (including anomalous topics) primarily manifest on a low-dimensional (keyword) subset of the full word space.  Thus, PTM is a natural clustering algorithm for identifying low-dimensional salient feature subsets for anomalous clusters.  The importance of this is supported by the experimental results in section 4.3, where use of LDA in place of PTM within ATD gave very poor results. Thus, we believe that the superior performance of our approach is the result of \textit{jointly} discovering candidate clusters and detecting their salient feature subsets;  accordingly, ATD's adapted use of PTM is a signature ``feature'' of the algorithm.

We also see that because the subspaces on which the anomalous topics manifest are very low dimensional, each document in the anomalous cluster, individually,  does not exhibit a high degree of atypicality. Only by considering all such anomalous documents \textit{collectively} can we detect the anomalous cluster. Individual point anomaly detection methods overall achieved low detection accuracy, especially on the real data sets.

\section{Conclusion}
\label{Concsection}
We have proposed an algorithm for detecting atypical topics exhibited by clusters of anomalous text documents. Unlike individual-based AD techniques, our method detects \textit{clusters} of anomalous documents which jointly manifest atypical topics on a small subset of (salient) features. Given a collection of normal documents, we first learn a (null) model for the typical topics. Then, in a separate test set batch, we detect all clusters of abnormal documents and the topics exhibited by them, one by one. We use statistical tests to determine the significance of any detected cluster. Our experiments show that our method can accurately detect anomalous topics and the subset of salient features under each such topic. Moreover, we show that, since only a small subset of words are salient in any anomalous topic, some standard AD methods, which evaluate atypicality on the full feature space, have low detection power. By contrast, our method accurately detects such anomalies by discovering salient feature subsets and detecting \textit{clusters} of anomalies.

\balance
\bibliographystyle{ieeetr}

\begin{thebibliography}{10}

\bibitem{HodgeVictoriaJ.2004}
V.~J. Hodge and J.~Austin, ``{A survey of outlier detection methodologies},''
  {\em Artificial Intelligence Review}, vol.~22, no.~2, pp.~85--126, 2004.

\bibitem{Chandola2009}
V.~Chandola, A.~Banerjee, and V.~Kumar, ``{Anomaly detection: A survey},'' {\em
  ACM Computing Surveys (CSUR)}, vol.~41, pp.~1--58, 2009.

\bibitem{Srivastava2008}
A.~Srivastava and A.~Kundu, ``{Credit card fraud detection using hidden Markov
  model},'' {\em IEEE Transactions on Dependable and Secure Computing}, vol.~5,
  no.~1, pp.~37--48, 2008.

\bibitem{Major2002}
J.~Major and D.~Riedinger, ``{EFD: A Hybrid Knowledge/Statistical-‐Based
  System for the Detection of Fraud},'' {\em Journal of Risk and Insurance},
  vol.~69, no.~3, pp.~309--324, 2002.

\bibitem{Wang2004}
K.~Wang and S.~Stolfo, ``{Anomalous payload-based network intrusion
  detection},'' in {\em Recent Advances in Intrusion Detection}, pp.~203--222,
  2004.

\bibitem{Kocak2014}
F.~Kocak, D.~Miller, and G.~Kesidis, ``{Detecting anomalous latent classes in a
  batch of network traffic flows},'' in {\em Information Sciences and Systems
  (CISS), 2014 48th Annual Conference on}, pp.~1--6, 2014.

\bibitem{Blei2003}
D.~M. Blei, A.~Y. Ng, and M.~I. Jordan, ``{Latent Dirichlet Allocation},'' {\em
  Journal of Machine Learning Research}, vol.~3, pp.~993--1022, 2003.

\bibitem{Blei2010a}
D.~Blei, L.~Carin, and D.~Dunson, ``{Probabilistic Topic Models},'' {\em
  Communications of the ACM}, vol.~55, pp.~77--84, Nov. 2012.

\bibitem{Soleimani2014}
H.~Soleimani and D.~J. Miller, ``{Parsimonious Topic Models with Salient Word
  Discovery},'' {\em Knowledge and Data Engineering, IEEE Transaction on},
  vol.~27, pp.~824--837, 2015.

\bibitem{Efron1979}
B.~Efron, ``{Bootstrap methods: another look at the jackknife},'' {\em The
  annals of Statistics}, pp.~1--26, 1979.

\bibitem{Xiong2011a}
L.~Xiong, s.~P. Barnab\'{a}, J.~G. Schneider, A.~Connolly, and V.~Jake,
  ``{Hierarchical probabilistic models for group anomaly detection},'' in {\em
  International Conference on Artificial Intelligence and Statistics},
  pp.~789--797, 2011.

\bibitem{Xiong2011}
L.~Xiong, B.~P\'{o}czos, and J.~Schneider, ``{Group anomaly detection using
  flexible genre models},'' in {\em Advances in neural information processing
  systems}, pp.~1071--1079, 2011.

\bibitem{Yu}
R.~Yu, X.~He, and Y.~Liu, ``{GLAD : Group Anomaly Detection in Social Media
  Analysis},'' in {\em Proceedings of the 20th ACM SIGKDD international
  conference on Knowledge discovery and data mining}, pp.~372--381, 2014.

\bibitem{Muandet2013}
K.~Muandet and B.~Sch\"{o}lkopf, ``{One-class support measure machines for
  group anomaly detection},'' in {\em 29th Conference on Uncertainty in
  Artificial Intelligence}, pp.~449--458, 2013.

\bibitem{Wong2002}
W.~Wong, A.~Moore, G.~Cooper, and M.~Wagner, ``{Rule-based anomaly pattern
  detection for detecting disease outbreaks},'' in {\em AAAI/IAAI}, pp.~217--223, 2002.

\bibitem{Wong2003a}
W.~Wong, A.~Moore, G.~Cooper, and M.~Wagner, ``{Bayesian network anomaly
  pattern detection for disease outbreaks},'' in {\em ICML}, pp.~808--815, 2003.

\bibitem{Das2008}
K.~Das, J.~Schneider, and D.~B. Neill, ``{Anomaly pattern detection in
  categorical datasets},'' in {\em KDD}, pp.~169--176, 2008.

\bibitem{McFowland2013}
E.~McFowland, S.~Speakman, and D.~Neill, ``{Fast generalized subset scan for
  anomalous pattern detection},'' {\em Journal of Machine Learning Research},
  vol.~14, no.~1, pp.~1533--1561, 2013.

\bibitem{Allan1998a}
J.~Allan, R.~Papka, and V.~Lavrenko, ``{On-line new event detection and
  tracking},'' in {\em SIGIR}, pp.~217--223, 1998.

\bibitem{Dai2010}
X.~Dai, Q.~Chen, X.~Wang, and J.~Xu, ``{Online topic detection and tracking of
  financial news based on hierarchical clustering},'' in {\em Machine Learning
  and Cybernetics (ICMLC), 2010 International Conference on}, pp.~3341--3346,
  2010.

\bibitem{He2010}
Q.~He, K.~Chang, E.-P. Lim, and A.~Banerjee, ``{Keep it simple with time: A
  reexamination of probabilistic topic detection models},'' {\em IEEE
  Transactions on Pattern Analysis and Machine Intelligence}, vol.~32, no.~10,
  pp.~1795--1808, 2010.

\bibitem{Schwarz1978}
G.~Schwarz, ``{Estimating the dimension of a model},'' {\em Annals of
  Statistics}, vol.~6, no.~2, pp.~461--464, 1978.

\bibitem{Dempster1977}
A.~P. Dempster, N.~M. Laird, and D.~B. Rubin, ``{Maximum likelihood from
  incomplete data via the EM algorithm},'' {\em Journal of the Royal
  Statistical Society.}, vol.~39, no.~1, pp.~1--38, 1977.

\bibitem{Meng1997}
X.-L. Meng and D.~{Van Dyk}, ``{The EM algorithm--an old folk-song sung to a
  fast new tune},'' {\em Journal of the Royal Statistical Society: Series B
  (Statistical Methodology)}, vol.~59, no.~3, pp.~511--567, 1997.

\bibitem{Pearson1900}
K.~Pearson, ``{On the criterion that a given system of deviations from the
  probable in the case of a correlated system of variables is such that it can
  be reasonably supposed to have arisen from random sampling},'' {\em The
  London, Edinburgh, and Dublin Philosophical Magazine and Journal of Science},
  vol.~50, pp.~157--175, 1900.

\bibitem{Agresti1996}
A.~Agresti, {\em {An introduction to categorical data analysis}}.
\newblock New York: Wiley, 1996.

\bibitem{Wilks1938}
S.~Wilks, ``{The large-sample distribution of the likelihood ratio for testing
  composite hypotheses},'' {\em The Annals of Mathematical Statistics}, vol.~9,
  no.~1, pp.~60--62, 1938.

\bibitem{Jordan1999}
M.~I. Jordan, Z.~Ghahramani, T.~S. Jaakkola, and L.~K. Saul, ``{An introduction
  to variational methods for graphical models},'' {\em Machine learning},
  vol.~37, no.~2, pp.~183--233, 1999.

\bibitem{Scholkopf2001}
B.~Sch\"{o}lkopf, J.~Platt, and J.~Shawe-Taylor, ``{Estimating the support of a
  high-dimensional distribution},'' {\em Neural computation}, vol.~13, no.~7,
  pp.~1443--1471, 2001.

\bibitem{Manevitz2001}
L.~M. Manevitz and M.~Yousef, ``{One-Class SVMs for Document Classification},''
  {\em Journal of Machine Learning Research}, vol.~2, pp.~139--154, 2001.

\bibitem{Zhao2009}
M.~Zhao and V.~Saligrama, ``{Anomaly Detection with Score functions based on
  Nearest Neighbor Graphs},'' in {\em Advances in neural information processing
  systems}, pp.~2250--2258, 2009.

\bibitem{Porter1980}
M.~Porter, ``{An algorithm for suffix stripping},'' {\em Program}, vol.~14,
  no.~3, pp.~130--137, 1980.

\end{thebibliography}

\end{document}